\documentclass[journal,10pt,twocolumn,twoside]{IEEEtran}
\usepackage{mathtools}
\usepackage{empheq}
\usepackage{xcolor}
\usepackage{algpseudocode, algorithm}
\usepackage{algorithmicx}
\usepackage{subfig}
\usepackage{varwidth}
\usepackage{url}
\usepackage{amssymb}
\usepackage{mathtools}
\usepackage{changes}
\usepackage[short]{optidef}
\usepackage{caption}
\usepackage{amsmath}
\usepackage{textcomp}
\usepackage{gensymb}
\usepackage{cite}
\usepackage{booktabs}
\usepackage{hyperref}
\usepackage{balance}
\usepackage{longtable}

\begin{document}

\title{Federated Learning for IoUT: Concepts, Applications, Challenges and Opportunities}

\author{Nancy Victor, Rajeswari. C, Mamoun Alazab, ~\IEEEmembership{Senior Member,~IEEE} Sweta Bhattacharya, Sindri Magnusson, Praveen Kumar Reddy Maddikunta, Kadiyala Ramana, Thippa Reddy Gadekallu, ~\IEEEmembership{Senior Member,~IEEE}
\thanks{Corresponding Author: Thippa Reddy Gadekallu}
\thanks{Nancy Victor, Rajeswari. C, Sweta Bhattacharya, Praveen Kumar Reddy Maddikunta, Thippa Reddy Gadekallu are with the School of Information Technology, Vellore Institute of Technology, Tamil Nadu-632014, India (Email: \{nancyvictor, rajeswari.c, sweta.b, praveenkumarreddy, thippareddy.g\}@vit.ac.in). }
\thanks{Mamoun Alazab is with College of Engineering, IT and Environment, and is the Director of the NT Academic Centre for Cyber Security and Innovation (ACCI) at Charles Darwin University, Australia. Email: alazab.m@ieee.org}
\thanks{Sindri Magnusson is with Department of Computer and Systems Science (DSV), Stockholm University, Borgarfjordsgatan, Sweden Email: sindri.magnusson@dsv.su.se}
\thanks{Kadiyala Ramana is with Department of Information Technology, Chaitanya Bharathi Institute of Technology, Hyderabad - 500075, Telangana, India. Email: ramana.it01@gmail.com}
}

\IEEEtitleabstractindextext{%
\begin{abstract}
\textcolor{black}{Internet of Underwater Things (IoUT) have gained rapid momentum over the past decade with applications spanning from environmental monitoring and exploration, defence applications, etc.} The traditional IoUT systems use machine learning (ML) approaches which cater the needs of reliability, efficiency and timeliness. However, an extensive review of the various studies conducted highlight the significance of data privacy and security in IoUT frameworks as a predominant factor in achieving desired outcomes in mission critical applications. \textcolor{black}{Federated learning (FL)} is a secured, decentralized framework  which is a recent development in machine learning,  that will help in fulfilling the challenges faced by conventional ML approaches in IoUT. This paper \textcolor{black}{presents an overview of the various applications of FL} in IoUT, its challenges, open issues and indicates direction of future research prospects. 
\end{abstract}


}

\maketitle

\IEEEdisplaynontitleabstractindextext

\IEEEpeerreviewmaketitle

\section{Introduction}
\begin{figure}[h!]
	\centering
	\includegraphics[width=\linewidth]{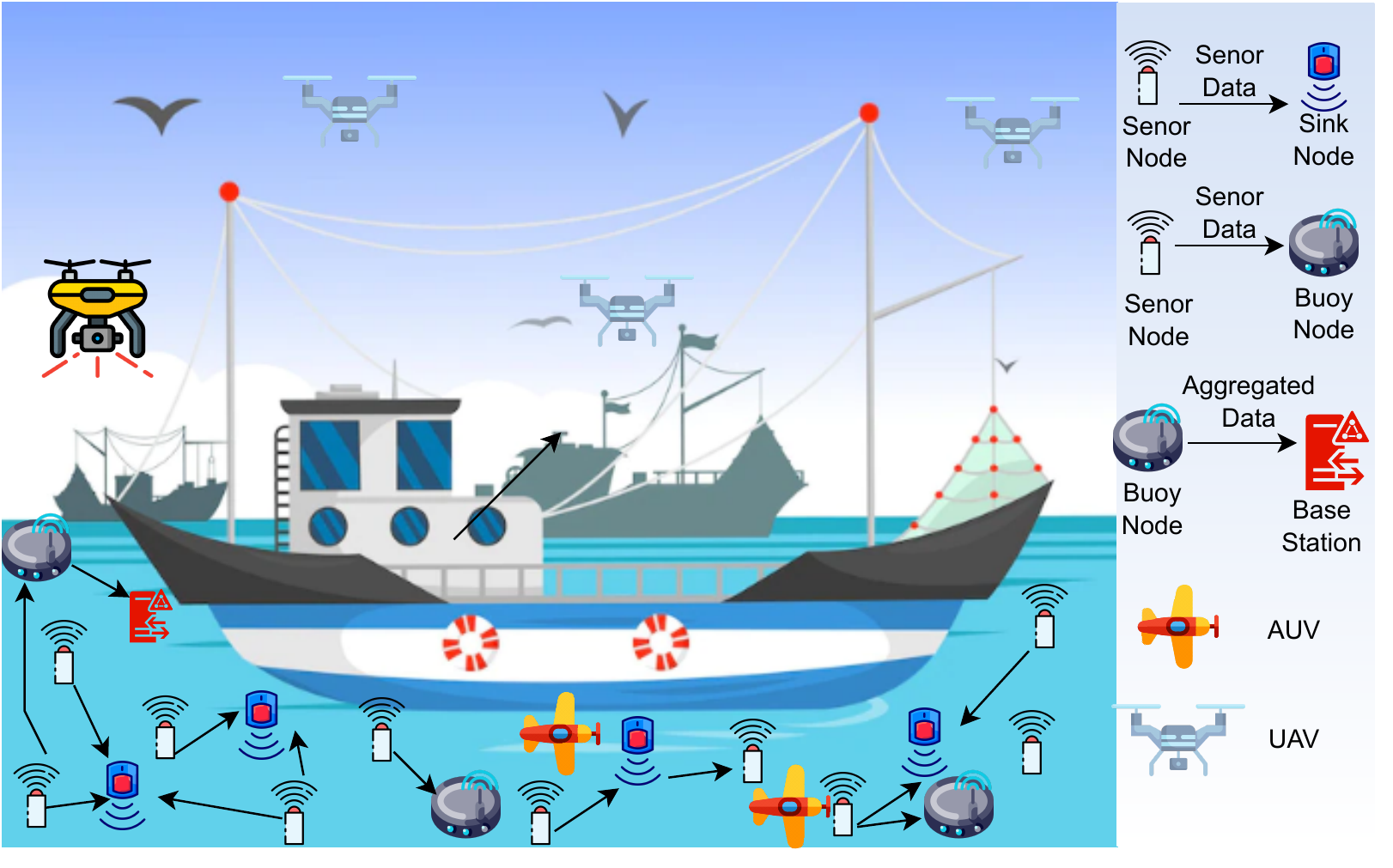}
	\caption{Network architecture of IoUT.}
	\label{fig_IoUT}
\end{figure}

Internet of Underwater Things (IoUT) is a network of \textcolor{black}{smart, interconnected underwater devices that help in monitoring the underwater environment and thereby using the data for various underwater applications such as environmental monitoring, disaster prediction and prevention, underwater exploration and so on.} IoUT has gained great momentum with the advances in \textcolor{black}{underwater sensor-based and underwater wireless communication-based technologies}. The underwater devices are embedded with sensors and extremely sophisticated tracking technologies, which enable capturing of data from the underwater environment and then reacts accordingly. The terrestrial objects like smart phones or similar devices get connected with underwater objects, which are further connected with virtual objects storing information pertaining to the features, origin and sensory aspect of the underwater objects. This information is extremely important for further analysis, tracking, monitoring and prediction, and thus is accessed by users with the help of Human-to-Thing and Thing-to-Thing form of communication. As an example, the underwater autonomous vehicles collect data, sense information, communicate with each other and then transmits information to control centers located on land for performing various types of predictive analytics. The information collected with the help of IoUT act as a source of conducting tasks relevant to crash surveying, searching of ship wrecks, prediction of tsunami, marine life health monitoring, aquatic data collection and various forms of archaeological expeditions. Fig.~\ref{fig_IoUT} depicts the network architecture of IoUT.

In the recent past, significant efforts have been dedicated in the development of practical IoUT applications catering to the need of fishing resource detection, environmental monitoring, defense and various others which all intend to add convenience and economic advantage to life on earth. The scope of marine exploration is ever evolving, which triggers the need of advanced mission critical IoUT systems. These systems help to fulfil non-functional requirements like latency, reliability, security, and timeliness. The inability to meet these requirements lead to inaccurate decision making that may cause disasters. Although IoUT based mission critical systems seem to have promising prospects, they have associated challenges as well. Firstly, information sensing often acts as a challenge in case of IoUT due to the vast span and complexity of underwater environment which damage or interfere with the data collection devices. This leads to inaccuracy and insufficiency in data collection affecting subsequent data analysis and the generated results. Secondly, transmission of information becomes extremely difficult due to variability in underwater temperature, hydrological characteristics, underwater pressure, varying water currents and topography interfering with the traditional communication models. Thirdly, the data collected from the underwater devices are often erroneous which includes large amount of irrelevant information creating delay in data transmission and generation of colossal waste of communication systems \cite{hou2021machine}. 
The traditional network and sensor based technologies employ deterministic methods which often fail to serve the purpose of IoUT based systems. Traditional network and sensor based technologies are classified into communication technology, transmission technology and energy harvesting technology. The communication technologies \textcolor{black}{include} IoUT based acoustic links. \textcolor{black}{Transmission technologies include} various RFID tags such as acoustic, radio and passive integrated transponder tags (PIT), and energy harvesting technologies such as microbial fuel cell (MFC), \textcolor{black}{that play} an important role in IoUT devices. The use of machine learning (ML) in IoUT based systems have revolutionized the traditional network methodologies fulfilling all the requirements of efficiency, fault tolerance, reliability, low-latency, information processing and information sensing. 

\textbf{Role of Federated Learning in IoUT Systems}:

The cloud computing based frameworks are immensely predominant in IoUT based applications. The data collected from the various devices are aggregated and sent to the cloud server. The required valuable information is retrieved from the large scale data set in the cloud server by applying ML algorithms. But they may fail to generate accurate results in case of mission critical IoUT systems due to high transmission latency. \textcolor{black}{The reason behind high latency is due to the longer distance between the cloud server and the IoUT devices, compelling it to be forwarded by multiple intermediate nodes.} Also, the collected data include lot of irrelevant information increasing wastage in resource consumption and enhanced latency. There is huge possibility of data getting compromised as well in this process impacting the security of the IoUT frameworks. Edge computing could be implemented in this regard. The autonomous underwater vehicles and similar floating devices which are in close proximity to the data sources could be considered as the edge nodes which would perform data processing at network edge eliminating useless data, reduce transmission data size and optimize utilization of resources. In addition to maximizing resource utilization and adding value to the data, the need of security is non-compromising. Federated Learning (FL) is a secured decentralized collaborative ML framework which has immense potential in developing efficient yet secured IoUT framework. As part of the FL based approach, each node in the network trains its subsequent sub-model independently. Each of these nodes interact only with the encrypted sub-model parameters in association with the fusion node to achieve an integrated global ML model \cite{alazab2021federated}. The FL based IoUT framework reduces communication overhead, and effectively mines from the large scale data set ensuring optimal security.

Even though there are several surveys on FL and IoUT separately, there is no survey on FL for IoUT to the best of our knowledge, \textcolor{black}{which} is a motivating factor for this survey. Some of the recent surveys on IoUT and FL are highlighted below and also summarized in Table I.

    In \cite{jahanbakht2021internet}, the authors have presented a comprehensive survey on the fusion of IoUT and big marine data analytics. In another interesting study, the authors in  \cite{jouhari2019underwater} surveyed the various networking techniques and communication technologies for IoUT. 
\begin{table*}[h!]
\centering
\caption{Summary of state-of-the-art implementation in IoUT.}
\label{tab:Summary}
\resizebox{\textwidth}{!}{%
\begin{tabular}{|l|p{7cm}|p{8cm}|}
\hline
Ref.No. &
  Contributions &
  Limitations \\ \hline
\cite{jahanbakht2021internet} &
  A review of IoUT network architecture, IoUT communication protocols, IoUT network topologies were presented. &
  This work focused mainly on IoUT but not FL and applications. \\ \hline
\cite{jouhari2019underwater} &
  Applications of IoT and ML for incentive designs in IoUT are studied. &
  This work did not focus on the potential of FL for IoUT and the use of IoT, AI techniques for IoUT. \\ \hline
\cite{boopalan2022fusion} &
  This work highlighted machine learning, deep learning and blockchain techniques for FL in secure IIoT. &
  This review paper was only limited to the IIoT with FL for achieving privacy and on device learning capability, applications of IIoT with FL in healthcare and automobile industry, while FL for IoUT were not studied. \\ \hline
\cite{pham2022aerial} &
 Applications of FL for aerial communications is presented. &
  This work focused mainly on FL for aerial communications  but not IoUT and applications. \\ \hline
\end{tabular}%
}
\end{table*}

The authors in \cite{boopalan2022fusion} have reviewed the applications of FL for industrial internet of things. In another interesting study, the authors in \cite{pham2022aerial} presented a survey of the applications of FL for aerial communications.

The rest of the paper is organized as follows. Section II gives an overview of IoUT and FL techniques. Section III discusses about the various applications of FL in IoUT and section IV explains about the various challenges, open issues and possible future directions for integration of FL with IoUT. Section V concludes the paper.
\section{Background}
In this section, a brief introduction about IoUT, FL and the motivation behind integration of FL with IoUT is discussed.
\subsection{Internet of Underwater Things}
\textcolor{black}{IoUT enable several autonomous vehicles in underwater to collaborate and communicate with each other. The sensors used in these vehicles can collect the data from the underwater environment that will further be transmitted to the storage such as cloud through the Internet. The information collected from the IoUT can be used for several purposes such as effective management of resources in the ocean, surveying shipwrecks and crashes, early detection of tsunamis/cyclones, monitoring the health of living species in the oceans, surveillance of marine environment, defense, and so on.} The IoUT can influence several applications from small scientific observatory, to a medium sized harbor to covering the ocean trade at global scale. Even though IoUT has many similarities with IoT with respect to the functioning and the structure, due to several factors like communication environment, computation/energy constrained resources, there are several differences between traditional IoT and IoUT \cite{jahanbakht2021internet}.

ML/deep learning techniques can be used to uncover the hidden patterns from the large volumes of data acquired from IoUT to realize several aforementioned applications. Some of the AI/ML based approaches in IoUT and their associated challenges are discussed below:

\textbf{AI in Information Sensing:} IoUT are usually deployed in harsh underwater environments wherein interruption and inaccurate information sensing is a prevalent issue. Also considering the energy consumption of these devices, it becomes extremely difficult to charge or replace the depleted battery, leading to inability of the systems to function properly. There are additional issues such as erroneous antenna signals, signal interferences, malicious attacks, and obstacles causing system failures and malfunction. The need for fault tolerant, reliable and effective information sensing in IoUT systems is thus a dire necessity. \textcolor{black}{ML and AI approaches help to establish relationship between the factors that contribute to device failures, which is difficult to be computed using a deterministic mathematical model. Deep learning techniques help in this regard by analysing large amount of historical data thereby generating interesting inferences and predictions. It also establishes  non-linear and dynamic relationship between probability of system failures and relevant underlying reasons in the complex underwater environment.} 
Multi resource data fusioning is a technique wherein diverse sensor data are integrated which eliminates lack of accuracy and robustness of collected data from single IoUT devices. The traditional data fusion strategies fail to provide the desired results due to its high-dimensional and non-linear nature. AI and machine learning approaches provide enhanced multi-sourced data fusion especially in \textcolor{black}{mission} critical IoUT systems. 

Challenges: The application of ML approaches in IoUT is dependent on data which often get compromised due to security concerns. The limited energy of the under water sensor networks (USWN) make the frameworks susceptible to malicious attacks reducing their lifespan. The nodes can be hacked by malicous attackers and the data collected by the nodes get read or manipulated resulting in generation of inaccurate decisions when AI algorithms are applied. 

\textbf{AI in Data Transmission}: \textcolor{black}{In IoUT}, there exist various challenges such as narrow availability of bandwidth, doppler effect, multi-path inferences in underwater wireless communication hindering efficient data transmission services. There are primarily four underwater wireless communication methods namely - \textcolor{black}{Under water Acoustic Communication (UAC)}, radio communication, wireless optical communication and magnetic - inductive communication. Instead of implementing these methods solely, the recent trend involves using multi-modal integrative communication methods which enables highly reliable transmission. The opto-acoustic communication method is one such approach which includes optical and acoustic communication to ensure higher rate and long distant data communication. The use of reinforcement learning play a significant role in this regard providing dynamic adaptive learning capability ensuring multi-dimensional complex channel perception and intelligent decision making. 

\textbf{Challenges}: Although the AI based multi-modal integrative communication methods have promising prospects, their high bit error rates, low bandwidth and propagation delays in the acoustic channel often invite malicious attacks. There are possibilities of routing attacks which prevent or divert the delivery of data packets in the network. Denial of service attacks are also possible which stop the legitimate nodes from performing their activities.

To summarize, conventional ML algorithms face several challenges in IoUT. Some of them are listed below:

\begin{itemize}
\item \textbf{Privacy Preservation:} The data collected from several autonomous vehicles such as submarines is very sensitive and should not be leaked into the wrong hands. In traditional ML approach, the collected data has to be transferred to the central cloud for further processing. This may result in exposing sensitive information and the malicious users may take advantage and get hold of this information while transmission or during storage.

\item \textbf{Increased Latency:} In conventional machine learning algorithms, data from all the devices/sensors has to be aggregated in the central cloud for training the machine learning algorithms. This incurs substantial communication cost.

\item \textbf{Handling Big Data:}
In IoUT, huge volumes of data will be generated from the devices/sensors at a rapid pace. Labeling of these data at real time and handling the volume and velocity of the data generated is a huge challenge.

\item \textbf{Quality of the Data:} Several data may be lost during the data transmission due to network issues. Also, external noise may be added to the data which poses a significant challenge regarding the quality of the data that has to be trained by the ML algorithms.
      
\end{itemize}

\subsection{Federated Learning} 

FL is a recent development of ML, where the global ML model is distributed across \textcolor{black}{the} individual devices. In FL, the data need not be transferred from individual devices to the central cloud storage, Instead, the ML algorithm itself will be executed in the local devices, and only the model updates will be transferred across the central storage for global ML training. As the raw data is not \textcolor{black}{sent across the} central server, the privacy of the data at local devices will not be exposed to the potential malicious users/devices. 

\subsection{Motivation for Integration of Federated Learning for Internet of Underwater Things}

The underlying principle of FL, where the  computation is offloaded to the local devices, has a great potential in solving several aforementioned challenges posed by training the data from IoUT through conventional ML approach. 

\begin{figure}[h!]
	\centering
	\includegraphics[width=\linewidth]{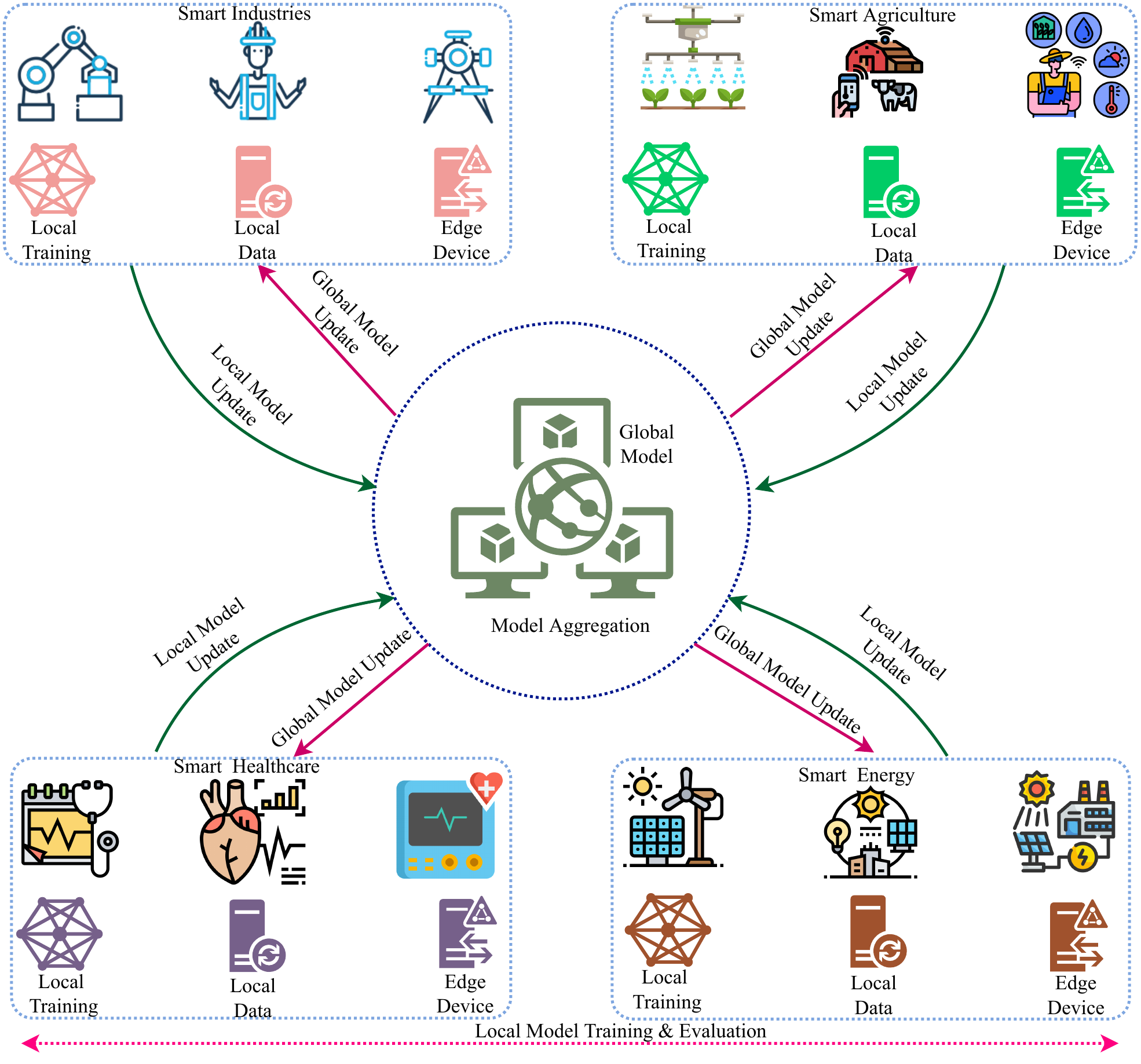}
	\caption{\textcolor{black}{Federated Learning for IoUT.}}
	\label{fig_FL}
\end{figure}

The motivation behind the integration of FL with IoUT are summarized as follows:
\begin{itemize}
    \item Preserving privacy of sensitive data generated from IoUT devices.  
    \item Reduced latency and communication costs while transmitting the data. 
    \item FL can handle the noisy data better when compared to traditional ML algorithms.
    \item Reduce the possibility of data quality issues during transmission of big data generated from the IoUT devices/sensors to the cloud.
\end{itemize}
The advantages of FL for IoUT are depicted in Fig. \ref{fig_FL}. 

\section{Applications}
In this section several applications of FL in different IoUT verticals are discussed.

\subsection{Environmental Monitoring}
    The water quality is monitored using a variety of parameters, including pH values, oxygen levels, temperature, and dissolved metals. A variety of drones and underwater robots are deployed to collect data and transmit it to the FL server. FL server provides timely and appropriate solutions, assisting engineers in making faster decisions and keeping polluted water away from the public. Several underwater sensors are deployed to monitor and detect oil spills at pipelines. The sensors installed outside and inside the pipelines measure various parameters of the oil, such as pressure, speed, and vibrations, and send the data to the sink node. The data from all sink nodes will be aggregated and sent to the base station (monitoring center). Based on the data gathered at the base station, a decision must be made as soon as possible to avoid disaster. Data transmission via an underwater physical transmission medium has a high propagation delay and error rate. To address these issues, the authors in \cite{gao2020federated} formed clusters from monitoring stations. Each cluster acquires local model training, and the central server consolidates all of the local model weight sets to create a global weight set with common features. The local cluster downloads the final global model for a new round of training resulting in faster decisions with higher accuracy. FL allows the devices to store and build the model in a decentralised fashion which is not possible in ML algorithms. The experimentation was carried out on i7-5930 processor with 12 cores at 3.50 GHz and 64 Gb of memory. \textcolor{black}{In the proposed model, more number of blocks could be transferred per second. Also,the encoding has been decreased by 9\% for 20 MB of files.}
\subsection{Underwater Exploration}
Underwater exploration aids in investigating the biological, chemical and physical conditions of the underwater environment so as to utilize it for scientific and commercial purposes. Technologies such as gears, satellites, underwater vehicles and buoys are generally used for exploring the underwater environment. Underwater exploration helps in discovering lost treasures and other natural resources, tracking underwater objects and also the fish density. Preserving data privacy, location privacy and device privacy is essential in such scenarios. FL approaches can significantly help in preserving the privacy of data collected by locally training the data, instead of sharing the data to a centralized server. This is a very unique feature for which FL is preferred over ML.

A "federated meta learning enhanced acoustic radio cooperative framework", also known as ARC/FML was proposed by H. Zhao et. al. \cite{zhao2021federated} to wisely use the data collected from distributed sources such as buoy nodes. This technique helps in sharing data across air and water. The ARC/FML technique can greatly help in sharing underwater exploration related sensitive data. The experiment was conducted using DeepSink and RF channel dataset. The proposed model acheives 97\% of accuracy. Similarly, an edge computing framework is proposed using blockchain and FL mechanisms by Z. Qin et. al. \cite{qin2021privacy}. This system can efficiently deal with the security issues in the marine IoT systems. The implementation suggests how malicious worker in FL can be simulated. During this process, initially the experimentation was carried with attack intensity of 20\% and achieved the loss function of FL to 0.092. Later the intensity rate was increased to 80\% and achieved loss function to 0.0195. As exploring underwater environment and gathering data regarding the same will be beneficial for the economy of the society, huge research prospects are awaiting in this field. 
\subsection{Disaster Prevention and Mitigation}
The underwater environment is always prone to disasters, both man-made and natural. Man-made disasters usually comprise of oil spills, poisonous gas and substance leakage and illegal dumping. One of the major man-made disasters happened in 2010, the Deepwater Horizon incident \cite{hagerty2010deepwater} on the "Oil Drilling Rig Deepwater Horizon". Concrete core was used to seal the well and nitrogen gas was used for curing. However, a gush of natural gas leaked through the core and it couldn't withstand the pressure, resulting in the gas to rise to the platform and catch fire. The explosion killed 11 workers and 94 members in the crew were cleared from the site. The rig was sunk and resulted in a catastrophic oil leakage from the well, and has has affected nearly 70000 square miles of ocean in the Gulf of Mexico. Natural disasters in ocean often fall into the categories such as tsunamis, hurricanes, and tropical storms. The Indian ocean earthquake and tsunami happened in 2004 is considered as one of the deadliest natural disasters \cite{telford2007international}.

FL enabled IoUT solutions significantly help in disaster prevention by monitoring the underwater environment in real-time. Data from the underwater environment can be collected using devices such as sensors, cameras, and underwater vehicles. These devices can be deployed at various locations, thus enabling continuous data collection in the underwater environment. Various sensors such as seismic pressure sensors can be used for disaster prevention. FL allows devices to learn shared prediction model rather storing it on the central server. Barrier systems are one of the effective ways in mitigating the effect of disasters. FL approaches are beneficial in disaster prevention not just because of preserving the data privacy, but also due to the improved model performance and scalability it offers.

\subsection{Defence/ Military} 
 
 \textcolor{black}{The naval activities in defence include }submarine detection, mine warfare, recovery operations, and surveillance. In the conventional model, these activities are carried out by \textcolor{black}{humans} in under water vessels. With advancement in technology, underwater wireless sensors (UWSN), a branch of wireless sensor networks, have proven to be a leading –edge communication infrastructure that enables under water activities without human intervention. UWSN serves to detect and classify subjects of interest in the underwater environment.

  
 The US navy emphasize their ships, submarines, drones and on-shore intelligence analysts to be able to share data in real time \cite{karagoz2012maritime}. Lack of strong data leveraging tools is a key challenge in the deployment of this Multi Domain Operations (MDO). AI and ML based approaches can address the overwhelming amount of data, improve data analytics for better decision making, increase the speed of action, reduced manpower and accommodate uncertainties also. Federated machine learning is a collaborative training approach where training data is not exchanged to the edge device so as to overcome constraints on training data sharing (policy, security, and coalition constraints) and insufficient network capacity. To address the aforementioned issues,  G. Cirincione and D. Verma \cite{cirincione2019federated} proposed a federated machine learning approach for deployment of MDO.

\subsection{Navigation and Localization}

Exploration of underwater environment requires modern assisted navigation and localization technologies. Some of the underwater sensors are anchored at the ocean bottom and some are floated at different positions or freely floated in the oceans current. The data collected from the various sensors are potentially useful only if the sensors' location is identified exactly. Also the underwater sensors location act as reference points for other smart objects, swimmers, divers and explorers. Location identification is the key point for source detection and tracking applications. Due to the characteristics of underwater channel, localization protocols cannot work in underwater application. Poor underwater positioning and navigation are the two predominant issues of IoUT. Sybil, black hole, and worm hole are significant attacks that utilizes or modifies the localization information generated by UWSNs. Improvements in the treatment of databases and model building could be adapted so as to improve the privacy in localization and navigation. Federated machine learning learn collaboratively to improve performance and overcome coalition sharing restrictions /network constraints that restrict sharing of training data. \textcolor{black}{Thus, a continuous FL process could be adapted for  localization and navigation, as the input is streaming data and the decision cycles are in seconds and milliseconds. This is another predominant reason why FL is preferred over ML in IoUT environments.}

\begin{figure}[h!]
	\centering
	\includegraphics[width=\linewidth]{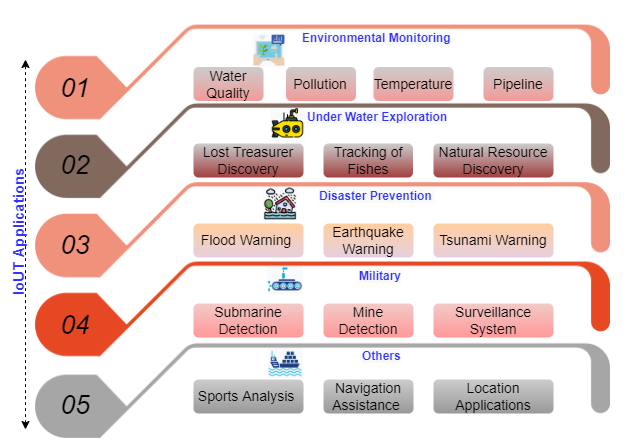}
	\caption{\textcolor{black}{Application verticals of IoUT.}}
	\label{fig_Applications}
\end{figure}

The applications of FL in IoUT are depicted in Fig. \ref{fig_Applications}.

\section{Challenges, Open Issues and Future Directions}
Integrating FL techniques in an IoUT environment proves to be advantageous due to the various reasons discussed in the previous sections. However, various challenges are also identified in such a setting. This section deals with the various challenges in integrating FL with IoUT, open issues for the research community and possible solutions that can be adapted. 

\textbf{Device/ Network Configuration:} Deploying FL over the network edge requires the configuration of devices or the network itself. Network configuration issues prominently occur in underwater networks due to the fact that the nodes are mostly mobile, thus resulting in disturbing the connectivity between nodes. This occurs due to various factors such as path loss, noise, multipath effect and doppler shift. Edge devices suffer from issues such as limited battery power, storage capacity, and computational capacity. In order to enable FL on IoUT, lightweight models with self/auto configuration methods need to be devised. Also, a hardware and algorithm co-design can significantly help. 

\textbf{Data Transmission:} Data transmission in underwater environment is different when compared with the terrestrial environment. One of the major issues is with respect to the low frequency range with which the signals need to be transmitted, so as to avoid the same being hindered by water. Due to the long transmission range, there is a high chance for interference and collision. When considering the deployment of FL in IoUT, this proves to be a significant challenge, as synchronous update is done at the server side, and this requires the data to be sent to the server even from the slowest client, in order to enable aggregation. However, due to the prevailing underwater environmental characteristics and the connectivity, availability and transmission issues, efficient solutions are required for enabling FL in IoUT. One of the promising research directions would be to use an asynchronous update at the server side \cite{sprague2018asynchronous}.  

\textbf{Unreliable Channel Condition:} Unreliable channel conditions can occur in underwater environment due to various factors such as limited bandwidth, channel noise, node mobility and transmission delays. Communication bottleneck is one of the significant challenges in deploying FL in an IoUT environment, that makes it difficult for the clients to share the local updates to a centralized server. Another bottleneck is in dealing with the dynamic changes that occur in IoUT networks since the topology itself can change over time due to the presence of mobile nodes.  Optimization techniques can be devised for deploying FL in an IoUT environment. Also, gradient compression schemes and aggregation frameworks that focus on computation, communication and resource efficiency can be used for dealing with bandwidth related issues. 
\begin{table*}[h!]
\centering
\caption{Challenges and the possible solutions for integrating FL in an IoUT environment.}
\label{tab:challenges}
\resizebox{\textwidth}{!}{%
\begin{tabular}{|p{4cm}|p{8.5cm}|p{4.5cm}|}
\hline
\textbf{Challenge   Type} &
  \textbf{Challenge   Description} &
 \textbf {Possible   Solutions and Effective Methods} \\ \hline
Device and   Network Configuration &
  Configuring the   devices and the network itself is challenging in an UIoT environment due to   the factors such as limited on-device resources, storage capacity and battery   power &
  Light weight models,   auto or self reconfiguration methods, hardware-algorithm co-design \\ \hline
Data   Transmission &
  Transmitting data in   UIoT suffers from significant issues such as network connectivity, data   availability, propagation speed, transmission range and rate, due to which   data aggregation is impeded at the server end &
  Asynchronous update   for FL \\ \hline
Unreliable   Channel Condition &
  Limited network   bandwidth, transmission delays, and noise are some of the factors that result   in an unreliable channel condition, that    will result in significant communication delays with respect to local   updates &
  Optimization   techniques, computation, communication and resource efficient aggregation   frameworks \\ \hline
System   Heterogeneity &
  The different types   of devices, data formats and software systems in an UIoT environment makes it   difficult for the FL technique to be integrated, since it can lead to   discrepancy in the local data structure in the clients' side &
  Frameworks that   support heterogeneity, global inference models \\ \hline
Privacy &
  Data, device and   location privacy need to be guaranteed to the clients in a FL-enabled UIoT   environment. Model inversion attacks result in the input data being accessed   by the attackers, thus questioning the client's privacy &
  Secure, light-weight   protocols \\ \hline
  Real-time generation of labels &
  For generating decisions in real-time, learning mechanisms need to applied, where real-time generation of labels is required, which is quite tedious in an underwater setting &
  Unsupervised learning mechanisms, Automated label generation \\ \hline
\end{tabular}%
}
\end{table*}
\textbf{System Heterogeneity:} The underlying hardware and software systems in an IoUT environment differ in various aspects when compared to the terrestrial IoT environment. This result in system design issues due to asynchronous communication, diverse data formats and dimensions. Enabling FL approaches in IoUT requires the handling of heterogenous devices and its capabilities, with the help of which load can be distributed among devices based on the availability. Frameworks that support system heterogeneity can be employed here, that will further enable in having a global reference model. 

\textbf{Privacy:} The sensor nodes in an IoUT environment are sparsely deployed, by making it quite difficult to manage. Hence, this is regarded as a complex and sensitive environment. Privacy, here deals with data, device and location privacy. Robust solutions are essential for preserving the privacy in IoUT networks. FL techniques help in dealing with these trust-related issues. However, FL also has privacy limitations that include reconstruction of user data by using the gradient information \cite{geiping2020inverting}. The images collected from the IoUT setting also can be reconstructed using attacks such as model inversion. This is indeed an open challenge to researchers to deal with the privacy related issues in a privacy-enabled setting. One possible solution could be the implementation of secure, light-weight protocols.

\textbf{Real-time generation of labels:} Generation of class labels is essential for applying supervised learning mechanisms in a dataset. However, in IoUT, generating labels in real-time is a time-consuming and tedious task. In order to address this issue, it is suggested to apply unsupervised learning mechanisms, where generation of class labels is not required. Another method would be to use automated tools for generating labels in real-time.

Table.\ref{tab:challenges} summarizes the challenges and the possible solutions for deploying FL in an IoUT environment. 

\section{Conclusion}
This paper presents a review of the IouT systems, its applications, challenges and scope of future direction of research. At the outset, a brief overview of IoUT and the underlying technologies have been discussed. The applications of AI/ML in information sensing and data transmission is presented. However the review of the various studies highlight the need of FL in achieving desired characteristics in an IoUT environment. The various application areas namely environment monitoring, defence, underwater exploration and disaster prevention is explored emphasizing on the integration of FL with IoUT. This reveals the potential underlying challenges of deploying FL in underwater environment and also highlights the possible solutions, effective methods indicating scope of future research.  

\bibliographystyle{IEEEtran}
\bibliography{Ref}
\end{document}